\documentclass[10pt,twocolumn,letterpaper]{article}
\usepackage[pagenumbers]{cvpr}
\usepackage{algorithm}
\usepackage{algpseudocode}
\definecolor{cvprblue}{rgb}{0.21,0.49,0.74}
\usepackage[pagebackref,breaklinks,hidelinks]{hyperref}
\hypersetup{pdftitle={WISER-RL: A Wake–Sleep Learning Mechanism for Robot Reinforcement Learning from One Demonstration},pdfauthor={Kwunhang Wong, Yang Bai, Qingyang Bao, Liudi Yang, Chenzhi Liu, Jiehong Lin, Wei Huang, Lin Li, Ziyuan Liu, Yujie Yang, Gitta Kutyniok, Zhongrui Wang, Xiaojuan Qi}}
\def\confName{CVPR}
\def\confYear{2027}
\newcommand{\wiserTitleLine}[1]{\raisebox{18pt}[\height][\depth]{#1}}
\title{\wiserTitleLine{WISER-RL: A Wake–Sleep Learning Mechanism for Robot} \\ \wiserTitleLine{Reinforcement Learning from One Demonstration}}
\newsavebox{\wiserAuthors}
\newcommand{\wiserByline}{%
\sbox{\wiserAuthors}{\parbox{\textwidth}{\centering    
\fontsize{12}{13}\selectfont

Kwunhang Wong$^{1,\ast}$\enspace
Yang Bai$^{2,3,\ast}$\enspace
Qingyang Bao$^{1}$\enspace
Liudi Yang$^{4}$\enspace
Chenzhi Liu$^{1}$\enspace
Jiehong Lin$^{1}$\\[2pt]
Wei Huang$^{1}$\enspace
Lin Li$^{5}$\enspace
Ziyuan Liu$^{6}$\enspace
Yujie Yang$^{1}$\enspace
Gitta Kutyniok$^{2,3}$\enspace
Zhongrui Wang$^{7,\dagger}$\enspace
Xiaojuan Qi$^{1,\dagger}$\par
\vspace{4pt}
\fontsize{8.5}{10}\selectfont
$^{1}$The University of Hong Kong \enspace
$^{2}$LMU Munich \enspace
$^{3}$Munich Center for Machine Learning (MCML) \enspace
$^{4}$University of Freiburg\\
$^{5}$HKUST \enspace
$^{6}$Huawei Heisenberg Research Center (Munich) \enspace
$^{7}$Southern University of Science and Technology\\[2pt]
$^{\ast}$Equal contribution \qquad
$^{\dagger}$Corresponding authors:
\href{mailto:wangzr@sustech.edu.cn}{\texttt{wangzr@sustech.edu.cn}} ;
\href{mailto:xjqi@eee.hku.hk}{\texttt{xjqi@eee.hku.hk}}\par}}%

\raisebox{\dimexpr(\dp\wiserAuthors-\ht\wiserAuthors)/2+8pt\relax}[0pt][0pt]{\usebox{\wiserAuthors}}}
\author{\wiserByline}
\begin{document}
\raggedbottom
\clubpenalty=10000
\widowpenalty=10000
\makeatletter
\g@addto@macro\@maketitle{%
  \begin{center}
    \makebox[\textwidth][c]{\includegraphics[width=0.95\textwidth]{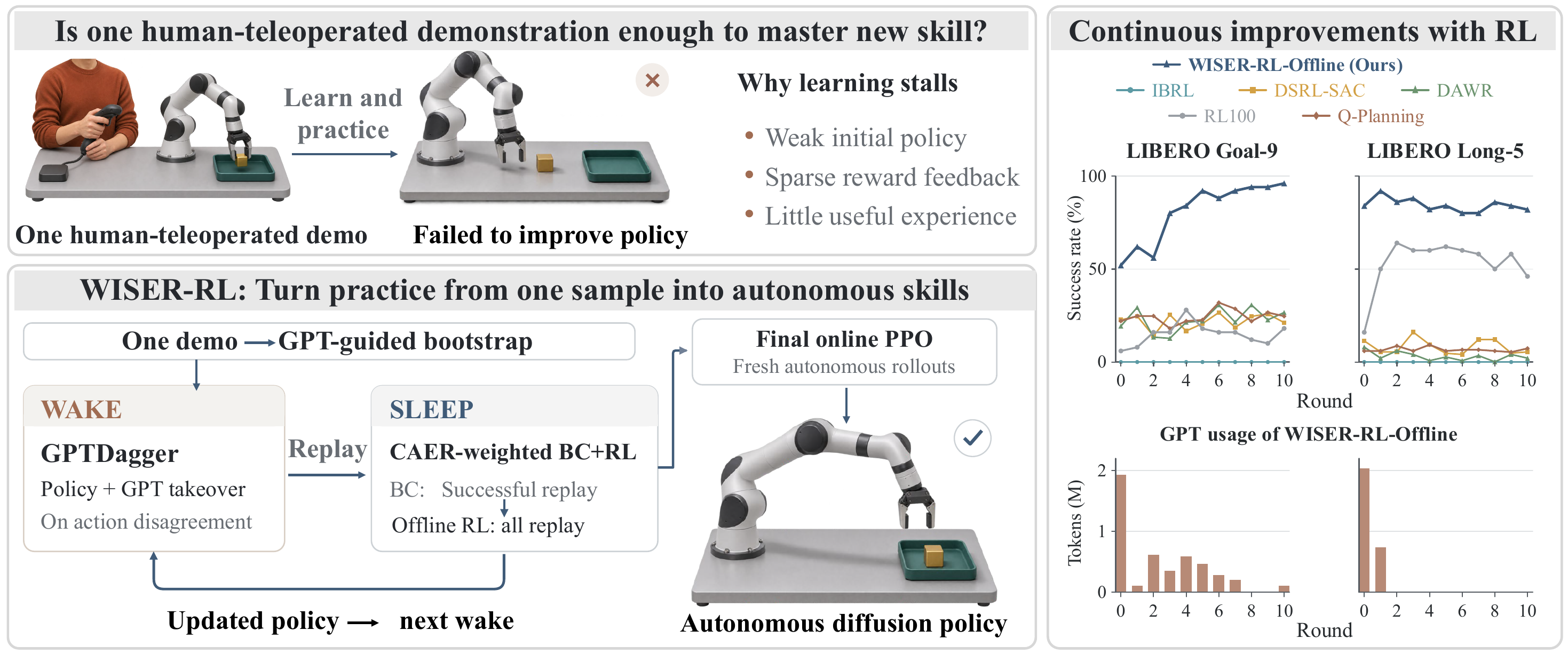}}
    \captionsetup{hypcap=false}
    \captionof{figure}{\textbf{WISER-RL overview.} Starting from one human demonstration, GPT-guided bootstrap initializes a diffusion policy. During wake, action disagreement triggers retrieval-guided GPT takeover to acquire experience; during sleep, CAER-weighted imitation and offline RL consolidate replay into the policy used in the next round. Final online PPO refines autonomous execution. Right: autonomous offline success versus five baselines on LIBERO Goal-9 and Long-5, together with per-round GPT token usage. Autonomous evaluation disables GPT and retrieval; online-refinement results are reported separately.}
    \label{fig:wiser_overview}
  \end{center}
}
\makeatother
\maketitle

\begin{abstract}
A robot learning from one human demonstration may rarely succeed on its own, leaving reinforcement learning with little reward feedback. We introduce WISER-RL (\textbf{W}ake--sleep policy improvement loop, \textbf{I}mitation, and \textbf{S}elective \textbf{E}xperience \textbf{R}eplay), a composite reinforcement learning framework that turns assisted practice into autonomous control. During wake, GPTDAgger monitors disagreement among actions sampled by a diffusion policy. When disagreement crosses a threshold, a frozen GPT controller uses retrieved experience to continue the episode. During sleep, Conflict-Aware Experience Reweighting (CAER) compares visually matched successful trajectories and downweights conflicting action windows with more steps remaining to success. The policy imitates the reweighted successes and learns from both successes and failures through offline advantage-guided RL before collecting the next round. After final online PPO refinement, the autonomous policy succeeded in 100\% evaluations on each of six LIBERO tasks and 100\% trials on each of two real-robot tasks.
\end{abstract}

\section{Introduction}
\label{sec:introduction}
A central goal of robotics is to acquire a new skill from one or a few human demonstrations and become proficient through practice~\cite{duan2017oneshot}. A person can watch an example, reason about how it applies to a different scene, and gradually perform the skill with less deliberate effort. For robots, this motivates combining rapid adaptation from examples with lasting autonomous competence.

Reinforcement learning (RL) refines robotic policies through interaction~\cite{ren2024dppo,lei2025rl100}, but initially weak policies struggle to collect useful experience. With sparse success rewards, many attempts can fail before the robot discovers a useful behavior~\cite{andrychowicz2017her,vecerik2017leveraging,nair2018overcoming,riedmiller2018playing,hu2023ibrl}. Recently, models such as GPT-6-Astra have demonstrated the ability to adapt from one or a few examples through in-context learning~\cite{cheng2026gptpolicy}. Drawing on strong reasoning and broad prior knowledge, these models can interpret a demonstration and apply it to a new scene. Yet this contextual ability does not automatically translate into a fast, efficient learned policy. \textbf{The challenge is to turn adaptation from a few examples into an autonomous policy that improves through practice.}

WISER-RL couples fast episodic adaptation with slow policy consolidation (Fig.~\ref{fig:wiser_overview}). A frozen contextual controller bootstraps a demonstration immediately, while a diffusion policy learns from the interaction rollouts across rounds.

One human demonstration can bootstrap GPT rollouts to initialize diffusion policy, but very costly~\cite{chi2023diffusionpolicy}. In \textit{wake} phase, GPTDAgger acts \textbf{decreasingly} when small policy's sampled actions acts improving. Retrieved experience and the executed prefix guide GPT intervention without training a task-specific teacher~\cite{xu2025rldg}. In \textit{sleep} phase, the policy imitates selected successful actions and uses offline RL to finetune from \textbf{both successes and failures}. The updated policy collects the next round of experience. Final online refinement uses fresh autonomous interaction for 100\% success.

A successful episode can contain a missed grasp and a recovery. When another episode reaches a visually similar state and grasps the object directly, both continuations receive the same terminal reward. Uniform imitation can therefore teach the retry alongside the direct grasp.

Inspired by adaptive forgetting, we propose \textbf{Conflict-Aware Experience Reweighting (CAER)} for reciprocal visual matches across successful episodes to modulate the sampling probability of action windows during behavioral cloning (BC). When matched action chunks conflict, CAER prunes imitating the window with more observed steps remaining to success. Offline RL can still learn from the outcomes of both trajectories. Our contributions are threefold:


\begin{enumerate}
 \item \textbf{Composite learning through WISER-RL.} We introduce an offline recurring wake--sleep mechanism that connects contextual experience acquisition with imitation and outcome-guided policy improvement, followed by final online refinement. Each consolidated policy drives subsequent interaction, providing a route from limited human demonstrations to autonomous control.
 \item \textbf{Conflict-aware consolidation through CAER.} We identify conflicting action replay memory across visually matched successful continuations and reweight imitation exposure according to observed efficiency, addressing direct imitate GPT policy unstable data source within the WISER-RL loop.
 \item \textbf{Simulation and real-robot evaluation.} Starting from one human demonstration per task, we demonstrate reducing GPT token cost and increasing autonomous success across training stages. After online refinement, the policy succeeded in 50/50 evaluations across six LIBERO tasks and 20/20 across two real-robot tasks. 
\end{enumerate}

\begin{figure*}[t]
  \centering
  \includegraphics[width=\textwidth]{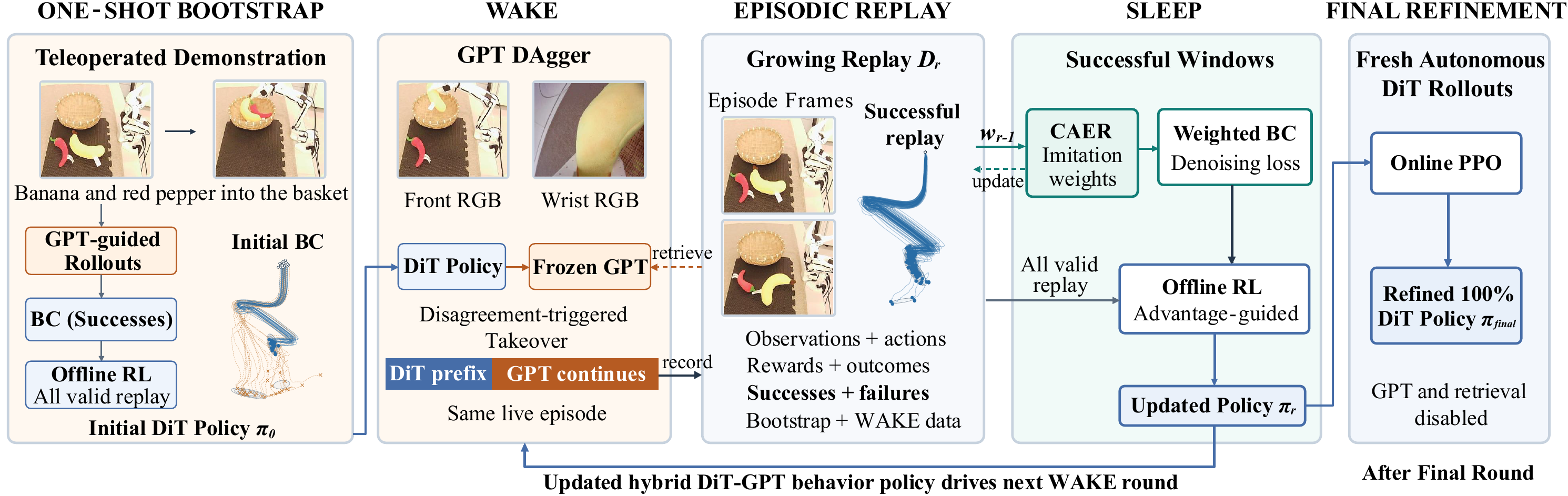}
  \caption{\textbf{WISER-RL framework.} one-shot bootstrapping initializes a diffusion policy, which improves through alternating wake--sleep rounds. During wake, diffusion-action disagreement triggers retrieval-guided GPT takeover, and rollouts are stored in episodic replay. During sleep, weighted behavior cloning and offline reinforcement learning update the policy. Final online refinement proceeds autonomously without GPT or retrieval. State-flow sketches are schematic.}
  \label{fig:wiser_framework}
  \vspace{-5pt}
\end{figure*}

\section{Related Work}
\label{sec:related}

\noindent\textbf{Offline-to-Online RL}.
Offline-to-online RL pretrains on both offline and freshly collected transitions in replay buffer to achieve safer, faster online-refinement~\cite{nair2020awac,nakamoto2023calql, lei2025rl100}. IBRL seeds its buffer with expert demonstrations and uses imitation-policy to guide exploration and value bootstrapping~\cite{hu2023ibrl}. Starting from a pretrained-vision-language-action (VLA) model by imitation, DSRL-SAC trains the frozen model in noise sampling space and expands replay buffer through rollout interaction~\cite{wagenmaker2025dsrl}; DAWR samples recent state--action pairs from replay buffer and weights each denoising loss by the action's advantage to update the diffusion policy~\cite{ren2024dppo}. Q-Planning learns a critic from rollouts to weight action proposals from a frozen pretrained-VLA~\cite{giridhar2026qplanning}. RL100 alternates imitation and offline policy improvement before final online refinement~\cite{lei2025rl100}. However, policy improvement can stall when weak policies rarely collect additional successful rollouts and receive little terminal reward feedback~\cite{andrychowicz2017her,nair2018overcoming,hu2023ibrl}. WISER-RL addresses this gap by coupling a hybrid behavioural policy GPTDAgger, followed by an advantage-guided RL in a wake--sleep loop.

\vspace{0.05in}\noindent\textbf{Emerging Training-free Policy Model}.
Traditional diffusion policies, VLAs, or world action models (WAMs) require training to adapt new robot control through imitation-based fine-tuning or retraining~\cite{chi2023diffusionpolicy,octo2024generalist, kim2024openvla,physicalintelligence2025pi05,ye2026dreamzero}. Prior works explore the in-context learning capability of existing VLA and WAM policies but have limited out-of-domain performance~\cite{fu2025icrt,sridhar2025ricl,macesanu2026victr}. Emerging GPT-Policy instead conditions a frozen GPT-6-Astra controller on demonstrations and shows promising training-free generalization across benchmarks~\cite{cheng2026gptpolicy}. Since repeated multimodal reasoning incurs substantial latency and token cost, WISER-RL consolidates this short-term contextual adaptation into long-term policy competence: fast episodic memory stores GPT-assisted experience, while selective imitation and RL distills the acquired behavior into a compact DiT policy for efficient autonomous control. 

\vspace{0.05in}\noindent\textbf{Memory Management and Consolidation}.
Experience replay enabled learning from stored interaction; complementary learning systems linked rapid episodic acquisition with gradual consolidation~\cite{lin1992replay,mcclelland1995complementary}. Later methods prioritized prediction error~\cite{schaul2016prioritized}, preserved representative experience~\cite{isele2018selective}, or revisited interference-prone samples~\cite{aljundi2019mir}. MemoryVLA retrieves and merges memories for action generation~\cite{shi2025memoryvla}. However, adaptive forgetting emphasizes reducing the influence of outdated or competing memories~\cite{richards2017persistence,anderson2021active}. In robot learning, earlier retries and later refinements may both succeed yet provide conflicting imitation targets. Inspired by this principle, WISER-RL uses CAER to reduce imitation of inefficient, conflicting experience while retaining it for retrieval and RL.

\section{WISER-RL: Continual Policy Improvement}
\label{sec:method}
\label{sec:objective}

\subsection{Preliminaries}
\noindent\textbf{Task Formulation.} For a fixed task instruction $\ell$, we model the environment as finite horizon Markov decision process (MDP) $\mathcal M=(\mathcal S,\mathcal A,P_0,P,r)$ with state space $\mathcal S$, continuous action space $\mathcal A\subseteq\mathbb R^d$, initial state distribution $P_0(s)\in\Delta(\mathcal{S})$ , transition kernel $P(s'\mid s,a):\mathcal S\times \mathcal A\to \Delta (\mathcal S)$, and terminal reward function $r(s,a):S\times \mathcal A\to \mathbb \{0,1\}$ under task horizon $T$ and discount factor $\gamma=1$. At step $t=0,\dots,T-1$, the policy $\pi_\theta$ conditions on the visual and proprioceptive observation $o_t$ and instruction $\ell$. Executing action $a_t$ induces $s_{t+1}\sim P(\cdot\mid s_t,a_t)$ and reward $r_t$. We consider one successful human demonstration $d_h$ for task $\ell$ forming the seed dataset $\mathcal D_h=\{d_h\}$ consisting success trajectory $d_h =(o_0,a_0,r_0,o_1\dots,o_L)$ of length $L\leq T$.

\vspace{0.05in}\noindent\textbf{Diffusion Policies.}
At each replanning step $t$, the diffusion policy $\pi_\theta(\cdot\mid o_t,\ell)$ predicts a normalized action chunk $X\in\mathbb R^{H\times d}$ spanning $H$ controls~\cite{chi2023diffusionpolicy}. For a training window starting at \(t\), we set \(x_0=X\), diffusion steps $k=1,\ldots,K$ and variances $\beta_k\in(0,1)$, the forward process is 
\begin{equation}
q(x_k\mid x_{k-1})=\mathcal N\!\left(\sqrt{1-\beta_k}\,x_{k-1},\,\beta_k I\right).  
\label{eq:diffusion}
\end{equation}
Writing $\bar\alpha_k=\prod_{j=1}^k(1-\beta_j)$ gives the marginal sample $x_k=\sqrt{\bar\alpha_k}\,X+\sqrt{1-\bar\alpha_k}\,\xi$ for noisy chunk, where $\xi\sim\mathcal N(0, I)$ follows standard Gaussian entries. For $k\sim\mathrm{Unif}\{1,\ldots,K\}$, the conditional denoiser $f_\theta$ predicts sampled noise $\xi$ by minimizing the per-window loss~\cite{ho2020denoising}
\begin{equation}
\mathcal L_{\mathrm{diff}}(\theta;o_t,X)
=
\mathbb E_{k,\xi}\!\left[
\left\|\xi-f_\theta(x_k,k,o_t,\ell)\right\|_2^2
\right].
\label{eq:diffusion_loss}
\end{equation}
BC minimizes the empirical average of this loss over successful trajectory windows under uniform weights. At inference, $x_K\sim\mathcal N(0,I)$ initializes the reverse process $x_{k-1}\sim p_{\theta,k}(\cdot\mid x_k,o_t,\ell)$ for $k=K,\ldots,1$. Its terminal sample $x_0$ specifies an action chunk. We adopt diffusion transformer (DiT) as diffusion policy backbone~\cite{chi2023diffusionpolicy} in the following.

\vspace{0.05in}\noindent\textbf{Learning Objective.}
Let $\pi_r=\pi_{\theta_r}$ be the policy after rollout interaction round $r$, and $\pi_0$ denote the initialized behavior policy bootstrapped on dataset $\mathcal D_h$. Under the initial state distribution $P_0$, the autonomous return at round $r$ is
\begin{equation}
\begin{aligned}
J(\pi_r)
&= \int_{\mathcal S} \mathbb E_{\pi_r,P}\!\left[
\sum_{t=0}^{T-1} r_t \,\middle|\, s_0=s
\right] P_0(\mathrm{d}s), \\
&\text{where }\; \pi_r^\star \in
\underset{\pi\in\{\pi_0,\ldots,\pi_r\}}{\arg\max}\; J(\pi).
\end{aligned}
\label{eq:policy_objective}
\end{equation}

Here $\pi_r^\star$ denotes the highest-return policy from bootstrap through round $r$. Round $r$ improves on the previous best when $J(\pi_r)>J(\pi_{r-1}^\star)$. As shown in Fig.~\ref{fig:wiser_framework}, WISER-RL uses $d_h$ as context for $N_{\mathrm{roll}}$ GPT rollouts and combines the valid trajectories with $d_h$ to form $\mathcal D_0$. Each subsequent round expands the replay buffer to form $\mathcal D_r$ during \textit{Wake} and updates the diffusion policy through biased imitation and unbiased advantage-guided offline RL during \textit{Sleep}. After the final wake--sleep round, WISER-RL refines $\pi^\star_R$ with online PPO using fresh autonomous rollouts.

\begin{figure*}[!t]
\centering
\includegraphics[width=0.92\textwidth]{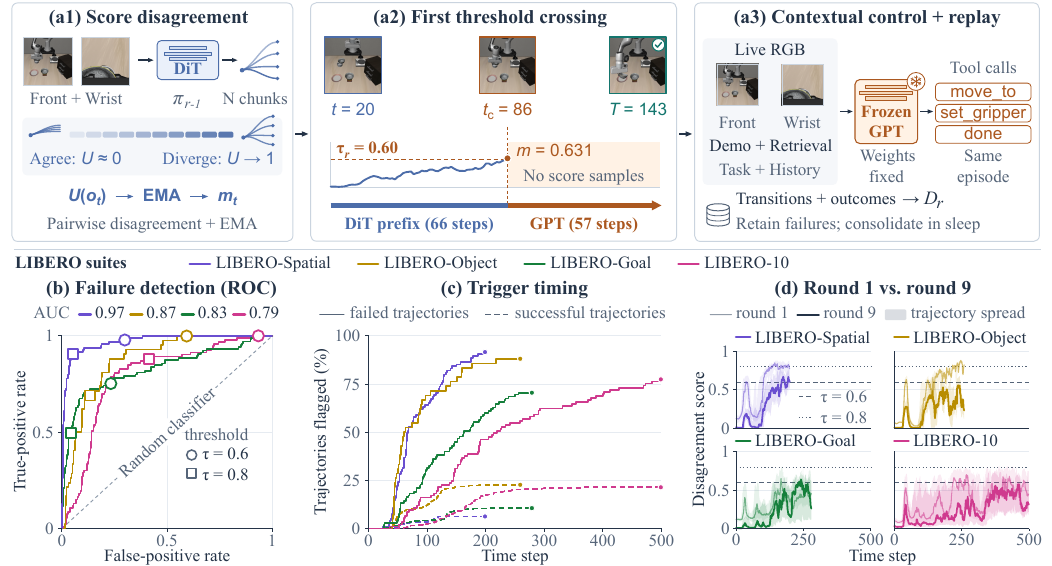}
\caption{\textbf{GPTDAgger procedure and diagnostics.} (a) Action-sample disagreement is smoothed by an exponential moving average (EMA); its first threshold crossing triggers GPT takeover in the same episode, and executed transitions enter replay. (b) Failure-detection ROC curves across four LIBERO suites; circles and squares mark thresholds $\tau=0.6$ and $0.8$. (c) Cumulative flagging of failed (solid) and successful (dashed) trajectories using suite-specific ROC-selected thresholds. (d) Median EMA disagreement in rounds 1 and 9, with interquartile bands. Horizontal lines mark thresholds $0.6$ and $0.8$.}
\label{fig:GPTDAgger}
\vspace{-5pt}
\end{figure*}

\subsection{Wake: In-Context Experience Acquisition}
\label{sec:wake}
DAgger~\cite{ross2011dagger} counters the compounding errors of BC by rolling out the learned policy, querying an expert at learner-visited states, and retraining on the aggregated data. Because labeling every visited state is costly, HG-DAgger have human intervene only when needed at the expert's own discretion~\cite{kelly2019hgdagger}. We propose \textbf{GPTDAgger} as a hybrid behaviour policy to keep human out of the loop in \textit{Wake} phase to guide uncertain DiT rollout towards completion. The resulting rollouts are aggregated in buffer for consolidation.

\vspace{0.05in}\noindent\textbf{Action Disagreement Trigger.} Given the observation $o_t$ and task instruction $\ell$ at round $r$, we draw $N\geq 2$ candidate chunks $X_t^{(1)},\dots,X_t^{(N)}\in\mathbb R^{H\times d}$ from the DiT chunk distribution $\pi=\pi_{r-1}(\cdot\mid o_t,\ell)$ by denoising independent Gaussian noise samples under the same conditioning. Using a Gaussian kernel $k_h(X,X')=\exp\bigl(-\lVert X-X'\rVert_F^2/(2h^2)\bigr)$ with bandwidth $h$ to measure the agreement of two chunks, we define the uncertainty of $\pi$ as the expected disagreement of two independent samples:
\begin{equation}
\label{eq:unc-def}
\begin{aligned}
\mathcal{V}_h(\pi)
  &\triangleq 1-\mathbb{E}_{X,X'\overset{\mathrm{iid}}\sim\pi}\bigl[k_h(X,X')\bigr]\\
  &= 1-\exp\Bigl(-\bigl[H_2(\pi*\mathcal{N}_\sigma)
       -H_2(\mathcal{N}_\sigma)\bigr]\Bigr),
\end{aligned}
\end{equation}
where $H_2(p)=-\log\int p^2$ denotes the R\'enyi quadratic entropy~\cite{renyi1961,principe2010}, and $\mathcal{N}_\sigma=\mathcal{N}(0,\sigma^2 I)$ with $\sigma=h/\sqrt{2}$. By Eq.~\eqref{eq:unc-def}, $\mathcal{V}_h(\pi)\in[0,1)$ grows monotonically with the entropy of the policy smoothed at scale $\sigma$, and it is zero iff the policy is deterministic at $(o_t,\ell)$. For a multimodal action distribution, this score captures both variation within each mode and separation between modes, as shown in the supplement. From these \(N\) candidates, we estimate $\mathcal{V}_h(\pi)$ with the U-statistic at the current observation~\cite{hoeffding1948}
\begin{equation}
\label{eq:wake_ustat}
{\mathcal{U}}(o_t)
  = 1-\frac{1}{N(N-1)}\sum_{i\neq j}
    \exp\!\left(-\frac{\lVert X_t^{(i)}-X_t^{(j)}\rVert_F^{2}}{2h^{2}}\right),
\end{equation}
The estimator in Eq.~\eqref{eq:wake_ustat} is unbiased for $\mathcal V_h(\pi)$ and has variance at most $1/(2N)$ regardless of the chunk dimension $H\cdot d$. To capture temporal uncertainty, let \(t_0,t_1,\ldots\) denote the scored steps. We initialize \(m_{t_0}=\mathcal U(o_{t_0})\) and update \(m_{t_j}=\rho\cdot m_{t_{j-1}}+(1-\rho)\cdot \mathcal U(o_{t_j})\) for \(j\geq1\). In the following, unless specified, we use $N=8$, $H=16$, $d=7$, $\rho=0.75$, and bandwidth $h=2$. Proofs and further properties are given in the supplementary material.

\vspace{0.05in}\noindent\textbf{GPTDAgger Policy.}
Much as human rely on practiced actions until an unfamiliar situation calls for deliberation, the \textit{Wake} mode DiT acts until $m_t>\tau_r$ for the first time; GPT then controls the remainder of the episode (Fig.~\ref{fig:GPTDAgger}; a1--a2). If the score never crosses, the rollout is fully autonomous and does not incur GPT tokens. At takeover, the frozen GPT issues a tool request $u_t\sim G_\phi(\cdot\mid o_t,\ell,\mathcal R_t,h_t)$ to controller by accepting multimodal inputs for in-context learning, where $u_t$ is a tool request, $\mathcal R_t$ is the retrieved experience, and $h_t$ is bounded history of executed prefix (Fig.~\ref{fig:GPTDAgger}; a3). $u_t$ is then translated into executable robot actions.

\begin{figure*}[!t]
\centering
\includegraphics[width=0.95\textwidth]{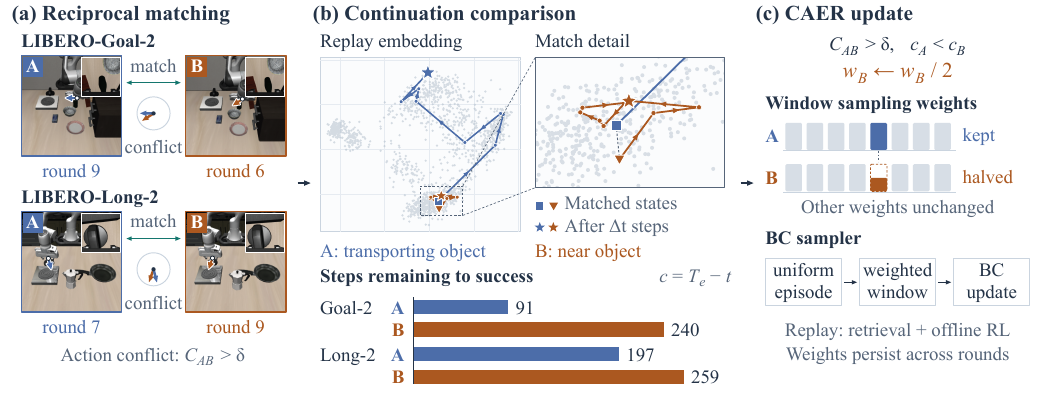}
\caption{\textbf{CAER motivation and update.} (a) Reciprocal matching identifies conflicting actions at similar observations in successful trajectories across rounds. (b) PCA paths of CLIP embeddings show $\Delta t=64$ steps after each match; stars mark intermediate states, distinct from termination $T_e$. Bars compare remaining steps to success. (c) CAER halves the slower window's BC sampling weight while preserving replay for retrieval and offline RL.}
\label{fig:caer_motivation}
\vspace{-1pt}
\end{figure*}

\vspace{0.05in}\noindent\textbf{Empirical Study of Trigger Threshold.}
We replayed autonomous LIBERO trajectories with their generating policies from $\mathcal D_0$ to $\mathcal D_{9}$ across 10 rounds. As shown in Fig.~\ref{fig:GPTDAgger}b, accounting episodes by their peak $m_t$ gave failure-flagging Area Under Curve (AUC) $0.79$--$0.97$ across four suites' ROC curves. At Fig.~\ref{fig:GPTDAgger}c diagnosis, we observe the action disagreement trigger can flag failed trajectories starting near $t=50$ in early-phase episodes, while later unconfident success can also be flagged with a flatter detection slope. The final Fig.~\ref{fig:GPTDAgger}d compares early-round and late-round disagreement triggers against the episode timestep. It shows the disagreement score reduces from roughly 0.8 to 0.6 as the DiT performance improves from round 1 to round 9; this establishes GPT token calls reduction as DiT improves.

\subsection{Sleep: Selective Policy Consolidation}
\label{sec:sleep}
A successful episode does not necessarily mean that all actions were perfectly executed. It may contain retries, detours, and conflicting local action targets. The success label only records the final outcome, while it could degrade quality for continual imitation in replay aggregation.

\vspace{0.05in}\noindent\textbf{Conflict-Aware Experience Reweighting.} We propose CAER to reweight observation--action windows from successful episodes in accumulated replay. At round $r$, a node $i=(e,t)$ indexes the window starting at step $t$ of success episode $e\in\mathcal D_r^+$, with front- and wrist-camera images and recorded action chunk $X_i$. A frozen visual CLIP encoder maps the images to $z_i$, the unit-normalized mean of the two camera features with $\|z_i\|_2=1$. For each node $i$, let $\mathcal E_i$ contain windows from every other successful episode in $\mathcal D_r^+$. Then set $\mathcal P_r$ contains the unordered reciprocal matches
\begin{equation}
 \begin{aligned}
 b(i)&=\arg\max_{j\in\mathcal E_i}z_i^\top z_j,\\
 \{i,j\}&\in\mathcal P_r\iff b(i)=j\ \text{and}\ b(j)=i.
 \end{aligned}
 \label{eq:buddy}
\end{equation}
Nodes with empty $\mathcal E_i$ remain unmatched; equal similarities are resolved by a fixed ordering of episode and window identifiers. Each unordered reciprocal pair is considered once per round. For its recorded normalized action chunks $X_i,X_j\in\mathbb R^{H\times d}$, we compute
\begin{equation}
 C_{ij}=\max(0, z_i^\top z_j) 
 \left[1-\exp\!\left(-\frac{\|(X_i-X_j)\|_F^2}{2h^2}\right)\right].
 \label{eq:buddy_conflict}
\end{equation}
Only pairs with $C_{ij}>\delta$, where $\delta$ is the conflict threshold, are treated as conflicting to reweight. This retrospective score compares stored continuations, whereas the \textit{Wake} score compares current policy samples for intervention.

Fig.~\ref{fig:caer_motivation}a illustrates the matched observations of Goal-2 and Long-2 as examples. For a matched node $i=(e,t)$ at step $t$ of episode $e$, let $T_e$ be that episode's successful termination step and $c_i=T_e-t$ the observed remaining steps to success. For each conflicting pair with $c_i<c_j$, CAER halves the weight of window $j$ (Fig.~\ref{fig:caer_motivation}b and c). Let $n_r(j)$ count the rounds through $r$ in which $j$ is the longer continuation of a conflicting pair. Each window enters replay with unit weight, so its entry in the weight vector $w_r$ of Algo.~\ref{alg:wiser_rl} is $w_r(j)=2^{-n_r(j)}$. A pair that qualifies in successive rounds progressively reduces $j$'s imitation exposure and \textbf{anchors towards} $i$. Equal remaining lengths leave both weights unchanged. BC then draws a successful episode $e$ uniformly, and draws its window start $t$ according to
\begin{equation}
 p_r(t\mid e)=\frac{2^{-n_r(e,t)}}
        {\sum_{u\in\mathcal W_e} 2^{-n_r(e,u)}},
 \label{eq:weighted_replay}
\end{equation}
where $\mathcal W_e$ is the set of eligible window start indices in episode $e$. The resulting BC replay distribution assigns window $(e,t)$ probability $q_r^+(e,t)=p_r(t\mid e)/|\mathcal D_r^+|$. CAER therefore changes local imitation exposure while preserving each episode's total sampling mass. Original experience remains available to retrieval and offline RL. 

\begin{algorithm}[t]
\caption{WISER-RL training}
\label{alg:wiser_rl}
\small
\begin{algorithmic}[1]
\Statex \textbf{Input:} Human seed $d_h$, initial policy $\pi_{\mathrm{init}}$, rollout size $N_{\mathrm{roll}}$, 
\Statex rounds $R$, intervention thresholds $\{\tau_r\}$.
\Statex \textbf{Bootstrap: seed and GPT experience}
\State $\mathcal B_0\gets\operatorname{GPTCollect}(d_h,N_{\mathrm{roll}})$
\State $\mathcal D_0\gets\mathcal D_h\cup\operatorname{Valid}(\mathcal B_0)$; $w_0\gets\mathbf 1$
\State $\pi_0^{\mathrm{BC}}\gets\operatorname{BC}(\pi_{\mathrm{init}},\mathcal D_0^+,w_0)$
\State $\pi_0\gets\operatorname{OfflineRL}(\pi_0^{\mathrm{BC}},\mathcal D_0)$
\For{$r=1,\ldots,R$}
    \Statex \hspace{\algorithmicindent}\textbf{Wake: GPT-assisted rollout collection}
    \State $\mathcal B_r\gets\operatorname{GPTDAgger}(\pi_{r-1},\mathcal D_{r-1},\tau_r,N_{\mathrm{roll}})$
    \State $\mathcal D_r\gets\mathcal D_{r-1}\cup\operatorname{Valid}(\mathcal B_r)$
    \Statex \hspace{\algorithmicindent}\textbf{Sleep: selective replay and consolidation}
    \State $w_r\gets\operatorname{CAER}(\mathcal D_r^+,w_{r-1})$
    \State $\pi_r^{\mathrm{BC}}\gets\operatorname{BC}(\pi_{r-1},\mathcal D_r^+,w_r)$
    \State $\pi_r\gets\operatorname{OfflineRL}(\pi_r^{\mathrm{BC}},\mathcal D_r)$
\EndFor
\Statex \textbf{Final online refinement: fresh autonomous interaction}
\State $\pi_{\mathrm{final}}\gets\operatorname{OnlinePPO}(\pi^\star_R)$
\State Evaluate $\pi_{\mathrm{final}}$ without GPT assistance
\Statex \textbf{Output:}  Refined policy $\pi_{\mathrm{final}}$
\end{algorithmic}
\end{algorithm}

\vspace{0.05in}\noindent\textbf{Imitation and offline reinforcement learning.}
At round $r$, diffusion policy initializes from $\pi_{r-1}$ and minimizes the per-window loss in Eq.~\ref{eq:diffusion_loss} under the CAER distribution $q_r^+$. For window $i=(e,t)$, let $o_i$ and $X_i$ denote its observation and recorded action chunk. The round-$r$ BC objective is
\begin{equation}
 \mathcal L_{\mathrm{BC},r}(\theta)=
 \mathbb E_{i\sim q_r^+}\big[\mathcal L_{\mathrm{diff}}(\theta;o_i,X_i)\big].
 \label{eq:bc}
\end{equation}
Offline RL initializes the actor from $\pi_r^{\mathrm{BC}}$ and fits twin critics $Q_{r,1},Q_{r,2}$ to bootstrapped return targets on $\mathcal D_r$, including failed transitions. The value function $V_r$ is fitted by IQL expectile regression~\cite{kostrikov2021iql,lei2025rl100}. The critics evaluate $a=X_{1:H'}$, the executed $H'$-action prefix of a normalized chunk, where $1\leq H'\leq H$; BC and disagreement use the full $H$-action chunk. The raw advantage is
\begin{equation}
 \widehat A_r(o,a)=\min_{i\in\{1,2\}}Q_{r,i}(o,a)-V_r(o).
 \label{eq:offline_advantage}
\end{equation}
The actor uses $\widetilde A_r=(\widehat A_r-\mu_B)/\max(\sigma_B,\epsilon_A)$, where $\mu_B,\sigma_B$ are the candidate minibatch mean and standard deviation and $\epsilon_A>0$ prevents division by zero. Let $q_r$ sample observations from valid replay, and let $x_{K:0}\sim p_{\mathrm{ref}}(\cdot\mid o)$ be a reference diffusion chain with $K$ denoising steps. The final sample $x_0$ is decoded and re-normalized to obtain $X$. At step $k$, define $\rho_k=p_\theta(x_{k-1}^{[H']}\mid x_k,o)/p_{\mathrm{ref}}(x_{k-1}^{[H']}\mid x_k,o)$, where $[H']$ selects the executed action coordinates. For each $k$, the actor maximizes
\begin{equation}
 \mathcal J_k=
 \mathbb E_{q_r p_{\mathrm{ref}}}\!\left[\min\!\left(\rho_k\widetilde A_r,
 \operatorname{clip}(\rho_k,1-\varepsilon,1+\varepsilon)\widetilde A_r\right)\right].
 \label{eq:offline_ppo}
\end{equation}
Here $\varepsilon>0$ is the clipping radius. The reference densities, sampled chain, and standardized advantages are fixed during its denoising updates; only the current policy is differentiated. This inherited advantage-guided improvement follows BPPO and RL100~\cite{zhuang2023behavior,lei2025rl100}. Imitation selects successful windows; offline RL uses successful and failed outcomes. These replay-based updates produce the policy for the next wake round. Algo.~\ref{alg:wiser_rl} summarizes the WISER-RL wake--sleep loop, including the check that quarantines unsafe GPT-assisted trajectories before they can enter replay at bootstrap ($\mathcal B_0$) or in any subsequent wake round ($\mathcal B_r$). 

\subsection{Autonomous Online Refinement}
\label{sec:online}
Final refinement starts from the last consolidated diffusion policy. With GPT takeover disabled, a fixed snapshot of the policy collects each batch of fresh rollouts. Diffusion PPO uses observed rewards and a learned value baseline to update the policy before the next batch~\cite{ren2024dppo,schulman2017ppo}. We evaluate the resulting policy autonomously with retrieval disabled.

\vspace{-2pt}
\section{Experimental Results}
\label{sec:experiments}
\subsection{Comparison Setup}
\noindent\textbf{Task and Evaluation.} For simulation tasks, we evaluate six LIBERO tasks: Goal-2, Goal-9, Spatial-5, Long-2, Long-5, and Long-8~\cite{liu2023libero}. The comparison baselines include IBRL, DSRL-SAC, DAWR, Q-Planning, and RL100. For real-world experiments, we evaluate WISER-RL with a 6-DoF SO-101 robot arm on two manipulation tasks: (1) “Stack green C block”: rotating light-green C-shaped block to align with dark-green C-shaped block and stacking it on top, testing precise alignment and placement; (2) “Banana and red pepper”: picking up banana and red pepper and placing both in the basket, testing long-horizon object grasping. The comparison baselines include RL100.

\vspace{0.05in}\noindent\textbf{Implementation Details.} For simulation tasks, WISER-RL, RL100, IBRL, DSRL-SAC, DAWR, and Q-Planning are all provided with 1 demonstration in round zero and continually expand 10 rollouts in each round, for a total of 10 rounds. Evaluation is performed on 50 official LIBERO states. For real-world tasks, WISER-RL bootstraps 20 demonstrations from 1 human demonstration in round zero, while RL100 is provided with 20 human demonstrations and continually expands 10 rollouts in each round, for a total of 5 rounds. We schedule WISER-RL $\tau_r$ linearly from 0.6 to 0.8 across $R$ rounds and report GPT tokens as a collection cost. Methods share each task's instruction, observation interface, and action representation. \textbf{RL100 serves as our no-GPT baseline}: it expands replay in simulation / real-world tasks with its own policy rollouts and uses the accumulated experience for imitation and offline RL. 

\begin{figure*}[t]
  \centering
  \includegraphics[width=0.88\textwidth]{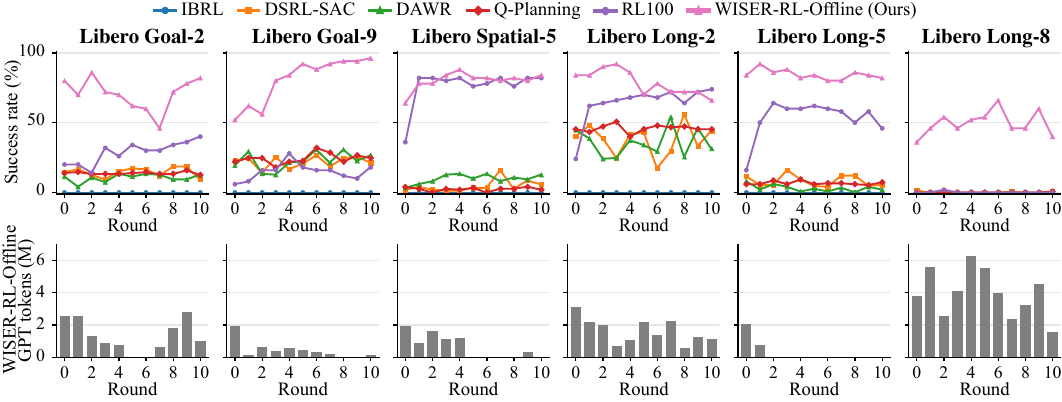}
  \vspace{-0.5em}
  \caption{\textbf{Offline learning on six LIBERO tasks.} Autonomous success (top) and per-round WISER-RL GPT token usage (bottom). For WISER-RL and GPTDAgger, round 0 reports the policy after initial offline RL.}
  \label{fig:libero_results}
  \vspace{-1em}
\end{figure*}

\begin{figure*}[t]
  \centering
  \includegraphics[width=0.88\textwidth]{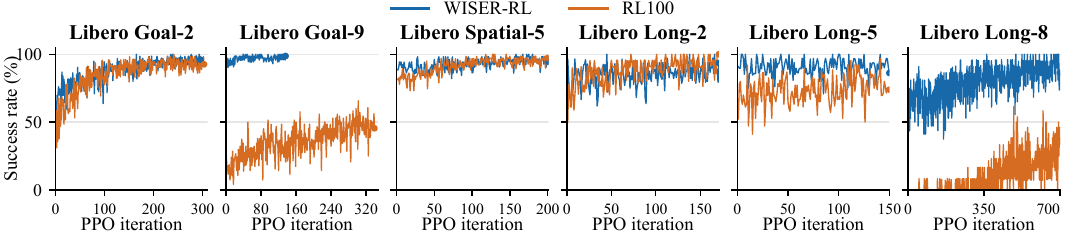}
  \vspace{-0.5em}
  \caption{\textbf{Online PPO refinement.} Autonomous training-rollout success of WISER-RL and RL100. Iteration ranges vary by task; separate evaluation results appear in Table~\ref{tab:stage_summary}.}
  \label{fig:ppo_results}
  \vspace{-1.5em}
\end{figure*}

\subsection{Offline Training Results}
Figure~\ref{fig:libero_results} compares autonomous success as replay grows. On Long-8, only WISER-RL regularly completes the task under this collection budget. On Goal-9, its lead grows over successive rounds. The comparison with RL100 is closer on Spatial-5, where RL100 catches up, and Long-2, where it finishes slightly ahead. The lower panels show the cost of acquiring experience under the linear increase in takeover threshold from 0.6 to 0.8. On Long-5, GPT usage is concentrated in the first two collections, and autonomous success remains high in later rounds. Long-8 still consumes tokens late in training while autonomous success fluctuates. Because the threshold changes across rounds and each takeover can span a different number of steps, falling token usage alone cannot establish policy consolidation. Autonomous evaluation supplies the complementary measure.

\begin{table}[t]
\centering\small
\setlength{\tabcolsep}{3pt}
\begin{tabular*}{\columnwidth}{@{\extracolsep{\fill}}lcccc@{}}
\toprule
& \multicolumn{4}{c}{\textbf{WISER-RL}}\\
\cmidrule(l){2-5}
& \textbf{Bootstrap} & \multicolumn{2}{c}{\textbf{Offline}} & \textbf{Online}\\
\cmidrule(lr){3-4}
\textbf{Task} & Initial BC & +GPTDAgger & +CAER & Refinement\\
\midrule
Goal-2 & 36/50 & 40/50 & 43/50 & \textbf{50/50}\\
Goal-9 & 26/50 & 45/50 & 48/50 & \textbf{50/50}\\
Spatial-5 & 39/50 & 45/50 & 44/50 & \textbf{50/50}\\
Long-2 & 41/50 & 44/50 & 46/50 & \textbf{50/50}\\
Long-5 & 39/50 & 46/50 & 46/50 & \textbf{50/50}\\
Long-8 & 14/50 & 28/50 & 33/50 & \textbf{50/50}\\
\bottomrule
\end{tabular*}
\caption{\textbf{WISER-RL on LIBERO.} Autonomous successes in 50 trials per task after bootstrap BC, at the best offline checkpoint for each variant, and after online PPO.}
\vspace{-14pt}
\label{tab:stage_summary}
\end{table}

\begin{figure}
\centering
{\includegraphics[width=1\columnwidth]{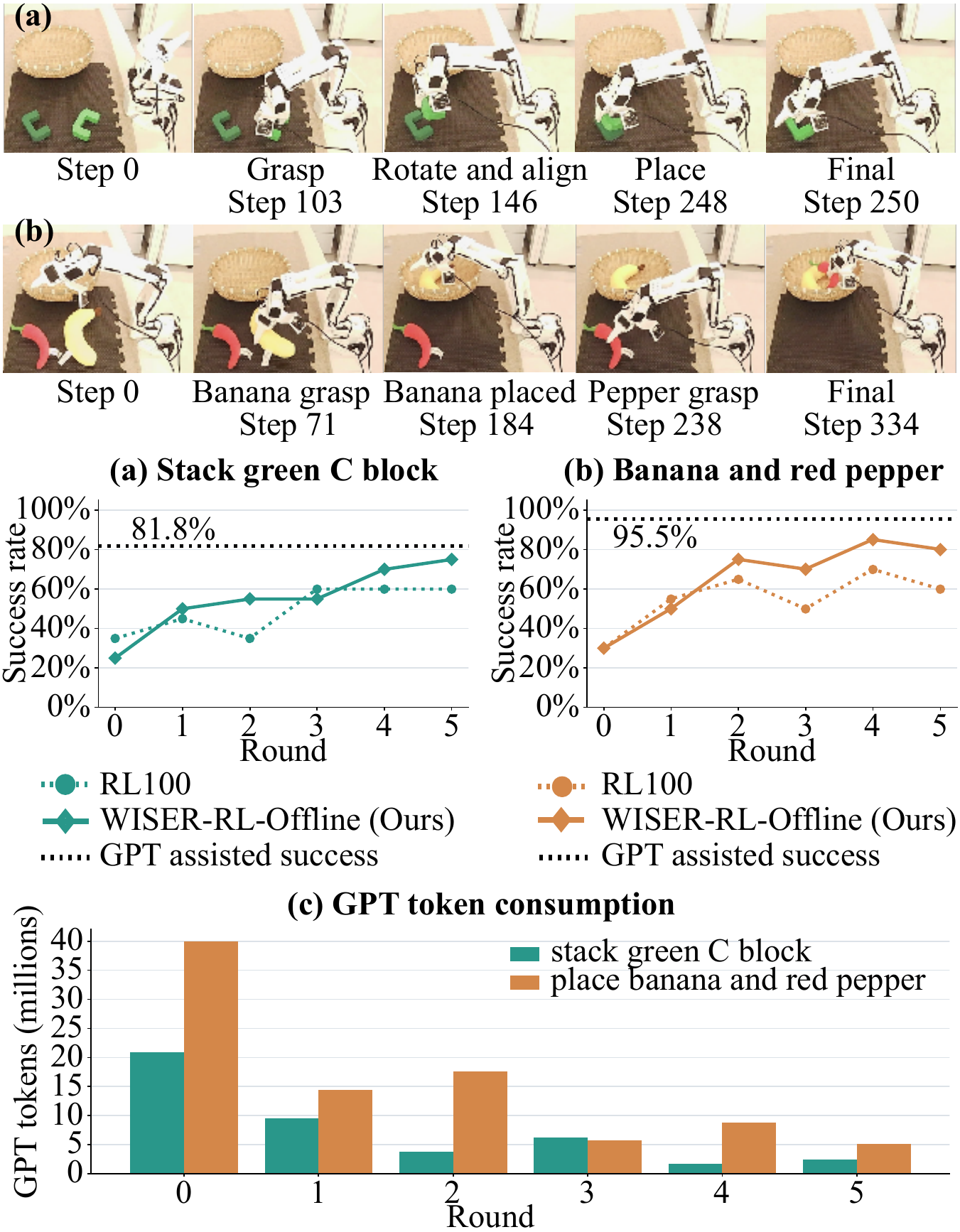}}
\caption{\textbf{Real-robot results.} Task keyframes, autonomous offline success (a,b), and GPT token usage (c). Dotted black lines denote GPT-assisted collection success; round 0 is bootstrap.}
\label{fig:real_robot}
\vspace{-14pt}
\end{figure}

Table~\ref{tab:stage_summary} separates initial BC from the subsequent offline updates. Round zero in Fig.~\ref{fig:libero_results} already includes offline RL. Summed over the six task evaluations, autonomous successes rise from 195/300 at bootstrap to 248/300 at the best GPTDAgger checkpoints and 260/300 with CAER. GPTDAgger improves on bootstrap BC on every task. CAER contributes a further gain on four tasks, with a tie on Long-5 (46/50) and a one-trial decrease on Spatial-5 (45/50 to 44/50). The largest CAER gain occurs on Long-8 (28/50 to 33/50). Figure~\ref{fig:caer_motivation} illustrates why a successful outcome alone can be ambiguous supervision. Two visually matched Goal-2 windows lead to success in 91 and 240 further steps. CAER halves the imitation weight of the slower conflicting window. Both trajectories still supply reward information to offline RL, so the same successful episode can be informative for value learning even where its actions provide a poor imitation target.

\subsection{Online PPO Refinement}
Figure~\ref{fig:libero_results} establishes how often the offline policy reaches the terminal reward before online updates begin. Figure~\ref{fig:ppo_results} then follows fresh autonomous PPO batches. On Goal-9 and Long-8, WISER-RL collects successful trajectories early, while RL100 initially receives fewer positive outcomes. With takeover disabled, every such trajectory supplies reward feedback on a complete sequence of actions chosen by the DiT itself. On Long-5, Goal-2, and Spatial-5, WISER-RL converge with slightly faster than RL100, so the GPT guided trajectories support policy improvement. Table~\ref{tab:stage_summary} reports the selected policies' autonomous evaluations. Section~4.1 specifies 50 official LIBERO states and three evaluation seeds. The table reports successes out of 50 per task, whereas each point in Fig.~\ref{fig:ppo_results} measures the policy collecting that training batch. All six WISER-RL policies reach 50/50 after refinement. Long-8 accounts for 17 of the 40 additional successes over the best offline checkpoints (33/50 to 50/50), compared with two on Goal-9 (48/50 to 50/50). Thus the same final score reflects GPT guided policy can consolidate completely into diffusion policy.

\subsection{Real-World Robot Results}
Figure~\ref{fig:real_robot} compares round-zero policies trained from two sources of supervision. WISER-RL uses only one human demonstration to bootstrap GPT collection of 20 more trajectories per task, while RL100 trains from 20 human demonstrations at start. The resulting autonomous policies tie on banana-and-pepper placement (6/20 each). On green-block stacking, which requires precise rotational alignment, WISER-RL starts two successes behind RL100 that uses offline human demonstrations (5/20 versus 7/20). Still, it shows GPT collection quality can be comparable to that of expert collection on diffusion policy BC.

The advantage emerges during replay expansion. Both methods collect ten rollouts per round for five rounds. RL100 adds the policy's own rollouts, while GPTDAgger allows a frozen vision-language controller to take over when the DiT's candidate actions disagree. By round five, the autonomous WISER-RL policy leads RL100 on stacking (15/20 versus 12/20) and on the two-object basket task (16/20 versus 12/20). Across these rounds, WISER-RL gains 50 percentage points on each task, compared with 25 and 30 points for RL100, respectively. 

The dotted lines report success during GPT-assisted collection (81.8\% for stacking and 95.5\% for basket placement), with the largest token cost at bootstrap. Table~\ref{tab:real_robot_results} reports autonomous evaluation at the best offline checkpoint: 15/20 for stacking and 17/20 for basket placement, which peaks in round four. Online PPO brings both tasks to 20/20.

\begin{table}[t]
\centering\small
\setlength{\tabcolsep}{3pt}
\begin{tabular}{@{}lccc@{}}
\toprule
\textbf{Task} & \textbf{Initial Bootstrap BC} & \textbf{Offline} & \textbf{Online}\\
\midrule
Stack green C block & 5/20 & 15/20 & \textbf{20/20}\\
Banana and red pepper & 6/20 & 17/20 & \textbf{20/20}\\
\bottomrule
\end{tabular}
\caption{\textbf{WISER-RL on the real-world robot.} Autonomous successes/20 after bootstrap BC, offline policy improvement, and online PPO refinement.}
\vspace{-1.5em}

\label{tab:real_robot_results}
\end{table}

\section{Conclusion}
We introduce WISER-RL as a composite RL framework that couples fast episodic memory with slow policy consolidation through a recurring wake--sleep policy-improvement loop. Contextual assistance acquires experience, selective imitation consolidates successful behavior, and offline advantage-guided RL uses task outcomes to improve the policy that drives subsequent collection. GPTDAgger and CAER provide the experience-acquisition and replay-reweighting mechanisms within this composition. Online PPO further improves the policy through fresh autonomous interaction and achieve faster convergence to 100\% success. 

\section{Limitation and Future Work}
In simulation, WISER-RL uses a frozen VLM to guide action selection, allowing the wake--sleep loop to collect experience and improve the policy without human intervention after the initial demonstration. On the physical robot, operators still reset the workspace between episodes and intervene when hardware or control failures interrupt execution. Future work should evaluate whether dedicated reset and recovery policies can reduce both the number of robot interactions and the amount of operator assistance required to reach a fully autonomous policy consolidation.

{\small\bibliographystyle{ieeenat_fullname}\bibliography{references}}

@inproceedings{fu2025icrt,
  title={{ICRT}: In-Context Imitation Learning via Next-Token Prediction},
  author={Fu, Letian and Huang, Huang and Datta, Gaurav and Chen, Lawrence Yunliang and Panitch, William Chung-Ho and Liu, Fangchen and Li, Hui and Goldberg, Ken},
  booktitle={IEEE International Conference on Robotics and Automation}, year={2025},
  doi={10.1109/ICRA55743.2025.11128272}
}

@inproceedings{sridhar2025ricl,
  title={{RICL}: Adding In-Context Adaptability to Pre-Trained Vision-Language-Action Models},
  author={Sridhar, Kaustubh and Dutta, Souradeep and Jayaraman, Dinesh and Lee, Insup},
  booktitle={Proceedings of the 9th Conference on Robot Learning}, pages={5022--5038}, year={2025}
}

@article{macesanu2026victr,
  title={{VICTR}: Value-Informed In-Context Learning for Vision-Language-Action Models},
  author={Macesanu, Luca and Zhang, Tianluo and Patil, Aditya and Sharma, Archit and Angers, Kevin and Clark, Christopher M. and Feng, Chen and Zhang, Juexiao},
  journal={arXiv preprint arXiv:2611.08698}, year={2026}
}

@inproceedings{ross2011dagger,
  title = {A Reduction of Imitation Learning and Structured Prediction to No-Regret Online Learning},
  author = {Ross, Stephane and Gordon, Geoffrey and Bagnell, Drew},
  booktitle = {Proceedings of the Fourteenth International Conference on Artificial Intelligence and Statistics},
  volume = {15},
  pages = {627--635},
  year = {2011},
  publisher = {PMLR},
  url = {https://proceedings.mlr.press/v15/ross11a.html}
}

@inproceedings{chi2023diffusionpolicy,
  title = {Diffusion Policy: Visuomotor Policy Learning via Action Diffusion},
  author = {Chi, Cheng and Feng, Siyuan and Du, Yilun and Xu, Zhenjia and Cousineau, Eric and Burchfiel, Benjamin and Song, Shuran},
  booktitle = {Robotics: Science and Systems},
  year = {2023},
  url = {https://diffusion-policy.cs.columbia.edu/}
}

@article{lei2025rl100,
  title = {{RL-100}: Performant Robotic Manipulation with Real-World Reinforcement Learning},
  author = {Lei, Kun and Li, Huanyu and Yu, Dongjie and Wei, Zhenyu and Guo, Lingxiao and Jiang, Zhennan and Wang, Ziyu and Liang, Shiyu and Xu, Huazhe},
  journal = {arXiv preprint arXiv:2510.14830},
  year = {2025},
  url = {https://arxiv.org/abs/2510.14830}
}

@article{kostrikov2021iql,
  title = {Offline Reinforcement Learning with Implicit {Q}-Learning},
  author = {Kostrikov, Ilya and Nair, Ashvin and Levine, Sergey},
  journal = {arXiv preprint arXiv:2110.06169},
  year = {2021},
  url = {https://arxiv.org/abs/2110.06169}
}

@article{cheng2026gptpolicy,
 title={In-Context Robot Learning with VLM Agents},
 author={Cheng, Dongzhou and Yi, Taoran and Fang, Ye and Zhang, Xingwu and Feng, Fan and Li, Yixuan and Zhuang, Gengxiong and Wang, Rongze and Yang, Shuai and Song, Wei and Xue, Weizhi and Wu, Minyan and Gui, Jie and Wang, Jiaqi and Wu, Tong},
 journal={arXiv preprint arXiv:2609.19138}, year={2026},
 url={https://arxiv.org/abs/2609.19138}
}

@article{ren2024dppo,
 title={Diffusion Policy Policy Optimization},
 author={Ren, Allen Z. and Lidard, Justin and Ankile, Lars L. and Simeonov, Anthony and Agrawal, Pulkit and Majumdar, Anirudha and Burchfiel, Benjamin and Dai, Hongkai and Simchowitz, Max},
 journal={arXiv preprint arXiv:2409.00588}, year={2024},
 url={https://arxiv.org/abs/2409.00588}
}

@article{schulman2017ppo,
 title={Proximal Policy Optimization Algorithms},
 author={Schulman, John and Wolski, Filip and Dhariwal, Prafulla and Radford, Alec and Klimov, Oleg},
 journal={arXiv preprint arXiv:1707.06347}, year={2017},
 url={https://arxiv.org/abs/1707.06347}
}

@article{shi2025memoryvla,
 title={{MemoryVLA}: Perceptual-Cognitive Memory in Vision-Language-Action Models for Robotic Manipulation},
 author={Shi, Hao and Xie, Bin and Liu, Yingfei and Sun, Lin and Liu, Fengrong and Wang, Tiancai and Zhou, Erjin and Fan, Haoqiang and Zhang, Xiangyu and Huang, Gao},
 journal={arXiv preprint arXiv:2508.19236}, year={2025},
 url={https://arxiv.org/abs/2508.19236}
}

@article{liu2023libero,
 title={{LIBERO}: Benchmarking Knowledge Transfer for Lifelong Robot Learning},
 author={Liu, Bo and Zhu, Yifeng and Gao, Chongkai and Feng, Yihao and Liu, Qiang and Zhu, Yuke and Stone, Peter},
 journal={arXiv preprint arXiv:2306.03310}, year={2023},
 url={https://arxiv.org/abs/2306.03310}
}

@inproceedings{zhuang2023behavior,
  title = {Behavior Proximal Policy Optimization},
  author = {Zhuang, Zifeng and Lei, Kun and Liu, Jinxin and Wang, Donglin and Guo, Yilang},
  booktitle = {International Conference on Learning Representations},
  year = {2023},
  url = {https://openreview.net/forum?id=3c13LptpIph}
}

@article{richards2017persistence,
  title={The Persistence and Transience of Memory},
  author={Richards, Blake A. and Frankland, Paul W.},
  journal={Neuron},
  volume={94}, number={6}, pages={1071--1084}, year={2017},
  doi={10.1016/j.neuron.2017.04.037}
}

@article{anderson2021active,
  title={Active Forgetting: Adaptation of Memory by Prefrontal Control},
  author={Anderson, Michael C. and Hulbert, Justin C.},
  journal={Annual Review of Psychology},
  volume={72}, pages={1--36}, year={2021},
  doi={10.1146/annurev-psych-072720-094140}
}

@inproceedings{andrychowicz2017her,
  title={Hindsight Experience Replay},
  author={Andrychowicz, Marcin and Wolski, Filip and Ray, Alex and Schneider, Jonas and Fong, Rachel and Welinder, Peter and McGrew, Bob and Tobin, Josh and Abbeel, Pieter and Zaremba, Wojciech},
  booktitle={Advances in Neural Information Processing Systems},
  year={2017},
  url={https://arxiv.org/abs/1707.01495}
}

@inproceedings{nair2018overcoming,
  title={Overcoming Exploration in Reinforcement Learning with Demonstrations},
  author={Nair, Ashvin and McGrew, Bob and Andrychowicz, Marcin and Zaremba, Wojciech and Abbeel, Pieter},
  booktitle={IEEE International Conference on Robotics and Automation},
  year={2018},
  url={https://arxiv.org/abs/1709.10089}
}

@inproceedings{kelly2019hgdagger,
  title = {{HG-DAgger}: Interactive Imitation Learning with Human Experts},
  author = {Kelly, Michael and Sidrane, Chelsea and Driggs-Campbell, Katherine and Kochenderfer, Mykel J.},
  booktitle = {IEEE International Conference on Robotics and Automation},
  year = {2019},
  url = {https://arxiv.org/abs/1810.02890}
}

@article{duan2017oneshot,
  title = {One-Shot Imitation Learning},
  author = {Duan, Yan and Andrychowicz, Marcin and Stadie, Bradly C. and Ho, Jonathan and Schneider, Jonas and Sutskever, Ilya and Abbeel, Pieter and Zaremba, Wojciech},
  journal = {arXiv preprint arXiv:1703.07326},
  year = {2017},
  url = {https://arxiv.org/abs/1703.07326}
}

@article{ye2026dreamzero,
  title={World Action Models are Zero-shot Policies},
  author={Ye, Seonghyeon and Ge, Yunhao and Zheng, Kaiyuan and Gao, Shenyuan and Yu, Sihyun and Kurian, George and Indupuru, Suneel and Tan, You Liang and Zhu, Chuning and Xiang, Jiannan and Malik, Ayaan and Lee, Kyungmin and Liang, William and Ranawaka, Nadun and Gu, Jiasheng and Xu, Yinzhen and Wang, Guanzhi and Hu, Fengyuan and Narayan, Avnish and Bjorck, Johan and Wang, Jing and Kim, Gwanghyun and Niu, Dantong and Zheng, Ruijie and Xie, Yuqi and Wu, Jimmy and Wang, Qi and Julian, Ryan and Xu, Danfei and Du, Yilun and Chebotar, Yevgen and Reed, Scott and Kautz, Jan and Zhu, Yuke and Fan, Linxi and Jang, Joel},
  journal={arXiv preprint arXiv:2602.15922},
  year={2026},
  url={https://arxiv.org/abs/2602.15922}
}

@article{lin1992replay,
  title={Self-Improving Reactive Agents Based on Reinforcement Learning, Planning and Teaching},
  author={Lin, Long-Ji},
  journal={Machine Learning}, volume={8}, pages={293--321}, year={1992},
  doi={10.1007/BF00992699}
}

@article{mcclelland1995complementary,
  title={Why There Are Complementary Learning Systems in the Hippocampus and Neocortex: Insights from the Successes and Failures of Connectionist Models of Learning and Memory},
  author={McClelland, James L. and McNaughton, Bruce L. and O'Reilly, Randall C.},
  journal={Psychological Review}, volume={102}, number={3}, pages={419--457}, year={1995},
  doi={10.1037/0033-295X.102.3.419}
}

@inproceedings{schaul2016prioritized,
  title={Prioritized Experience Replay},
  author={Schaul, Tom and Quan, John and Antonoglou, Ioannis and Silver, David},
  booktitle={International Conference on Learning Representations}, year={2016},
  url={https://arxiv.org/abs/1511.05952}
}

@inproceedings{isele2018selective,
  title={Selective Experience Replay for Lifelong Learning},
  author={Isele, David and Cosgun, Akansel},
  booktitle={Proceedings of the AAAI Conference on Artificial Intelligence}, volume={32}, number={1}, year={2018},
  doi={10.1609/aaai.v32i1.11595}
}

@inproceedings{aljundi2019mir,
  title={Online Continual Learning with Maximally Interfered Retrieval},
  author={Aljundi, Rahaf and Caccia, Lucas and Belilovsky, Eugene and Caccia, Massimo and Lin, Min and Charlin, Laurent and Tuytelaars, Tinne},
  booktitle={Advances in Neural Information Processing Systems}, year={2019},
  url={https://arxiv.org/abs/1908.04742}
}

@article{nair2020awac,
  title={{AWAC}: Accelerating Online Reinforcement Learning with Offline Datasets},
  author={Nair, Ashvin and Gupta, Abhishek and Dalal, Murtaza and Levine, Sergey},
  journal={arXiv preprint arXiv:2006.09359}, year={2020},
  url={https://arxiv.org/abs/2006.09359}
}

@article{hu2023ibrl,
  title={Imitation Bootstrapped Reinforcement Learning},
  author={Hu, Hengyuan and Mirchandani, Suvir and Sadigh, Dorsa},
  journal={arXiv preprint arXiv:2311.02198}, year={2023},
  url={https://arxiv.org/abs/2311.02198}
}

@article{wagenmaker2025dsrl,
  title={Steering Your Diffusion Policy with Latent Space Reinforcement Learning},
  author={Wagenmaker, Andrew and Nakamoto, Mitsuhiko and Zhang, Yunchu and Park, Seohong and Yagoub, Waleed and Nagabandi, Anusha and Gupta, Abhishek and Levine, Sergey},
  journal={arXiv preprint arXiv:2506.15799}, year={2025},
  url={https://arxiv.org/abs/2506.15799}
}

@article{giridhar2026qplanning,
  title={Beyond Imitation: Self-Improving Robot Policies via Off-Policy Q-Planning},
  author={Giridhar, Varun and Khandelwal, Anant and Collins, Jeremy A. and Georgiev, Ignat and Garg, Animesh},
  journal={arXiv preprint arXiv:2608.21204}, year={2026},
  url={https://arxiv.org/abs/2608.21204}
}

@inproceedings{nakamoto2023calql,
  title={{Cal-QL}: Calibrated Offline {RL} Pre-Training for Efficient Online Fine-Tuning},
  author={Nakamoto, Mitsuhiko and Zhai, Yuexiang and Singh, Anikait and Mark, Max Sobol and Ma, Yi and Finn, Chelsea and Kumar, Aviral and Levine, Sergey},
  booktitle={Advances in Neural Information Processing Systems}, year={2023},
  url={https://arxiv.org/abs/2303.05479}
}

@article{octo2024generalist,
  title={Octo: An Open-Source Generalist Robot Policy},
  author={{Octo Model Team} and Ghosh, Dibya and Walke, Homer and Pertsch, Karl and Black, Kevin and Mees, Oier and Dasari, Sudeep and Hejna, Joey and Kreiman, Tobias and Xu, Charles and Luo, Jianlan and Tan, You Liang and Chen, Lawrence Yunliang and Sanketi, Pannag and Vuong, Quan and Xiao, Ted and Sadigh, Dorsa and Finn, Chelsea and Levine, Sergey},
  journal={arXiv preprint arXiv:2405.12213}, year={2024},
  url={https://arxiv.org/abs/2405.12213}
}

@article{kim2024openvla,
  title={{OpenVLA}: An Open-Source Vision-Language-Action Model},
  author={Kim, Moo Jin and Pertsch, Karl and Karamcheti, Siddharth and Xiao, Ted and Balakrishna, Ashwin and Nair, Suraj and Rafailov, Rafael and Foster, Ethan and Lam, Grace and Sanketi, Pannag and Vuong, Quan and Kollar, Thomas and Burchfiel, Benjamin and Tedrake, Russ and Sadigh, Dorsa and Levine, Sergey and Liang, Percy and Finn, Chelsea},
  journal={arXiv preprint arXiv:2406.09246}, year={2024},
  url={https://arxiv.org/abs/2406.09246}
}

@article{physicalintelligence2025pi05,
  title={$\pi_{0.5}$: A Vision-Language-Action Model with Open-World Generalization},
  author={{Physical Intelligence} and Black, Kevin and Brown, Noah and Darpinian, James and Dhabalia, Karan and Driess, Danny and Esmail, Adnan and Equi, Michael and Finn, Chelsea and Fusai, Niccolo and Galliker, Manuel Y. and Ghosh, Dibya and Groom, Lachy and Hausman, Karol and Ichter, Brian and Jakubczak, Szymon and Jones, Tim and Ke, Liyiming and LeBlanc, Devin and Levine, Sergey and Li-Bell, Adrian and Mothukuri, Mohith and Nair, Suraj and Pertsch, Karl and Ren, Allen Z. and Shi, Lucy Xiaoyang and Smith, Laura and Springenberg, Jost Tobias and Stachowicz, Kyle and Tanner, James and Vuong, Quan and Walke, Homer and Walling, Anna and Wang, Haohuan and Yu, Lili and Zhilinsky, Ury},
  journal={arXiv preprint arXiv:2504.16054}, year={2025},
  url={https://arxiv.org/abs/2504.16054}
}

@inproceedings{xu2025rldg,
  author    = {Xu, Charles and Li, Qiyang and Luo, Jianlan and Levine, Sergey},
  title     = {{RLDG}: Robotic Generalist Policy Distillation via Reinforcement Learning},
  booktitle = {Proceedings of Robotics: Science and Systems},
  year      = {2025},
  doi       = {10.15607/RSS.2025.XXI.028}
}

@inproceedings{ho2020denoising,
  title     = {Denoising Diffusion Probabilistic Models},
  author    = {Ho, Jonathan and Jain, Ajay and Abbeel, Pieter},
  booktitle = {Advances in Neural Information Processing Systems},
  volume    = {33},
  pages     = {6840--6851},
  year      = {2020}
}

@article{vecerik2017leveraging,
  author  = {Vecerik, Mel and Hester, Todd and Scholz, Jonathan and Wang, Fumin and Pietquin, Olivier and Piot, Bilal and Heess, Nicolas and Roth{\"o}rl, Thomas and Lampe, Thomas and Riedmiller, Martin},
  title   = {Leveraging Demonstrations for Deep Reinforcement Learning on Robotics Problems with Sparse Rewards},
  journal = {arXiv preprint arXiv:1707.08817},
  year    = {2017}
}

@inproceedings{riedmiller2018playing,
  author    = {Riedmiller, Martin and Hafner, Roland and Lampe, Thomas and Neunert, Michael and Degrave, Jonas and van de Wiele, Tom and Mnih, Vlad and Heess, Nicolas and Springenberg, Jost Tobias},
  title     = {Learning by Playing -- Solving Sparse Reward Tasks from Scratch},
  booktitle = {Proceedings of the 35th International Conference on Machine Learning},
  volume    = {80},
  pages     = {4344--4353},
  year      = {2018}
}

@inproceedings{renyi1961,
  author = {R{\'e}nyi, Alfr{\'e}d},
  title = {On Measures of Entropy and Information},
  booktitle = {Proceedings of the Fourth Berkeley Symposium on Mathematical Statistics and Probability},
  volume = {1}, pages = {547--561}, year = {1961},
  publisher = {University of California Press}
}

@book{principe2010,
  author = {Pr{\'i}ncipe, Jos{\'e} C.},
  title = {Information Theoretic Learning: {Renyi's} Entropy and Kernel Perspectives},
  publisher = {Springer}, year = {2010}, doi = {10.1007/978-1-4419-1570-2}
}

@article{hoeffding1948,
  author = {Hoeffding, Wassily},
  title = {A Class of Statistics with Asymptotically Normal Distribution},
  journal = {Annals of Mathematical Statistics}, volume = {19},
  number = {3}, pages = {293--325}, year = {1948}
}
\end{document}